\def\year{2019}\relax
\documentclass[letterpaper]{article} 
\usepackage{aaai19}  
\usepackage{times}  
\usepackage{helvet}  
\usepackage[usenames,dvipsnames]{color}
\usepackage{courier}  
\usepackage{url}  
\usepackage{graphicx}  
\usepackage{listings}

\usepackage{balance}

\usepackage[utf8]{inputenc}
\usepackage[english]{babel}
\usepackage{paralist}
\usepackage{amssymb,amsthm}
\theoremstyle{definition}
\newtheorem{definition}{Definition}[section]

\newcommand{\xqed}[1]{%
    \leavevmode\unskip\penalty9999 \hbox{}\nobreak\hfill
    \quad\hbox{\ensuremath{#1}}}
\newcommand{\Endofdef}{\xqed{\lozenge}}

\graphicspath{{./images/}}

\usepackage{sgame}
\usepackage{multirow}

\usepackage{comment}

\usepackage{tikz}
\usetikzlibrary{arrows,automata}
\usepackage{pgfplots}
\usepackage{pgfplotstable}
\pgfplotsset{compat=1.3}
\usetikzlibrary{pgfplots.dateplot}
\usepackage{filecontents}
\usetikzlibrary{backgrounds}
\usepgfplotslibrary{fillbetween}

\graphicspath{{images/}}

\usepackage{enumitem}

\usepackage{subcaption}

\usepackage{algorithm}
\usepackage{algcompatible}
\usepackage{algpseudocode}

\usepackage{tikz}
\usepackage{pgfplots}
\usepackage{tikz-qtree}

\title{Markov Game Modeling of Moving Target Defense for Strategic Detection of Threats in Cloud Networks}
\author{Ankur Chowdhary$^*$, Sailik Sengupta$^*$, Dijiang Huang, Subbarao Kambhampati\\
    Computing, Informatics, and Decision Systems Engineering\\
    Arizona State University\\
}
\begin{document}
    \maketitle
    \begin{abstract}
        The processing and storage of critical data in large-scale cloud networks necessitate the need for scalable security solutions.  It has been shown that deploying all possible security measures incurs a cost on performance by using up valuable computing and networking resources which are the primary selling points for cloud service providers.  Thus, there has been a recent interest in developing Moving Target Defense (MTD) mechanisms that helps one optimize the joint objective of maximizing security while ensuring that the impact on performance is minimized.
        Often, these techniques model the problem of multi-stage attacks by stealthy adversaries as a single-step attack detection game using graph connectivity measures as a heuristic to measure performance, thereby (1) losing out on valuable information that is inherently present in graph-theoretic models designed for large cloud networks, and (2) coming up with certain strategies that have asymmetric impacts on performance.
        In this work, we leverage knowledge in attack graphs of a cloud network in formulating a zero-sum Markov Game and use the Common Vulnerability Scoring System (CVSS) to come up with meaningful utility values for this game. Then, we show that the optimal strategy of placing detecting mechanisms against an adversary is equivalent to computing the mixed Min-max Equilibrium of the Markov Game. We compare the gains obtained by using our method to other techniques presently used in cloud network security, thereby showing its effectiveness. Finally, we highlight how the method was used for a small real-world cloud system.
    \end{abstract}
    
    \begin{figure}[!t]
        \centering
        \includegraphics[width=0.48\textwidth]{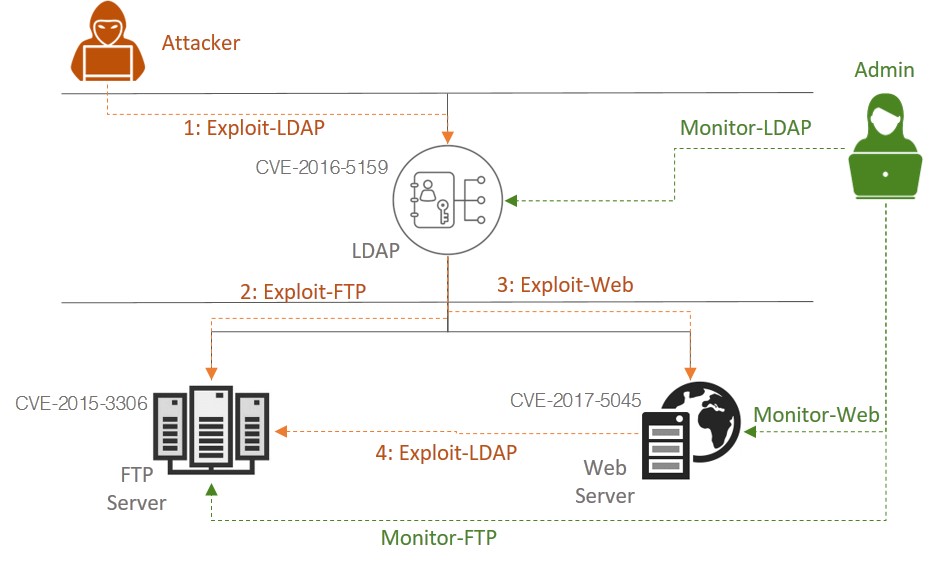}
        \vspace{-2em}
        \caption{An example cloud network scenario in which an attacker can take various attack paths and the defender wants to find a strategy for placing intrusion detection systems.}
        \label{fig:scenario_1}
        \vspace{-0.8em}
    \end{figure}
    
    \section{Introduction}
    A cloud service provider provides processing and storage hardware along with networking resources to customers for profit. Although a cloud provider might want to use state-of-the-art security protocols, vulnerabilities in software desired or run by customers can put sensitive information stored in or communicated over the cloud at risk.
    
    Distributed elements such as firewalls, Intrusion Detection Systems (IDS), log monitoring systems etc. have been the backbone to detect (or stop) malicious traffic entering such systems. Unfortunately, the scale of modern-day cloud systems makes the placement of all possible detecting and monitoring mechanisms an expensive solution \cite{jha2002two,venkatesan2016moving,senguptamoving}, using up the computing and network resources of the cloud that could have been better utilized by giving to customers which in turn would be better for business. Thus, the question of how one should place a limited number of detection mechanisms to limit the impact on performance while ensuring that the security of the system is not drastically reduced becomes a significant one.
    
    There has been an effort to answer this question in previous research works \cite{venkatesan2016moving,senguptamoving}. Researchers have pointed out that a static placement of detection systems is guaranteed to be insecure because an attacker, with reconnaissance on their side (which is there by design), will eventually learn this static placement and thereby avoid it. Thus, dynamic placement of these detection mechanisms has become a default. Such an approach is popularly known as Moving Target Defense (MTD) and can be used for shifting the detection surface where the set of attacks monitored changes in some randomized way after every time step, thereby introducing uncertainty if an attack will be caught. Although one can vary the length of this time step to introduce further complexity, we assume, similar to a majority of work done in the MTD community, this time step is fixed and decided beforehand.
    
    Previous work often treats the cloud system in a way similar to that of a physical security system where the primary challenge is to allocate a limited set of security recourses (IDS) to an asset/schedule (network/host) that needs to be protected \cite{dobss,sinha2015physical}.
    In the case of cloud-systems, a global dynamic allocation strategy has two major problems.  First, the treatment of multi-step attacks as single-step attacks as individual and independent attacks leads to the sub-optimal placement of detection mechanisms because they are inherently myopic failing to effectively leverage the information present in the system design. Such strategies, for example, may prioritize detection a high-impact attack on a web-server more than a low-impact attack on a path that leads an attack to a storage server, which when exploited may have major consequences. Second, these methods can come up with strategies where pure strategies where multiple detection systems are placed on the same host have high non-zero probabilities, leading to degradation of performance (for a customer situated) on that host more than others.
    
    In this paper, we try to address these problems by modeling the cloud system as a Markov Game. A sub-networks in the cloud network, determined using the system's Attack Graph (AG), represents the states of our game. The attacker actions correspond to real-world attacks determined using the Common Vulnerabilities and Exploits (CVEs) found in the National Vulnerability Database (NVD) and the defender actions correspond to the placement of detection systems that can detect these (known) attacks. We design the rewards of this game by leveraging the inherent security knowledge present in the Common Vulnerability Scoring Systems (CVSS). This helps us design defender strategies that take into account the long-term impacts of multi-stage attacks while ensuring that the defender picks a limited number of monitoring actions in each state of the game. The latter ensures the placement of detection mechanisms that do not affect the performance asymmetrically in the different parts of the network. The key contributions of this research work are:
    \begin{itemize}
        \item We design an attack graph based multi-stage attack analysis method leveraging Markov Game Modeling to optimize the cost incurred and the security provided by detection mechanisms in a multi-tenant cloud network. 
        \item We show that the Markov Game strategy for the placement of IDS performs better than other static and randomization strategies--the improvement margin widens when the size of the sub-net modeled in a state increases in size and the discount factor of the game approaches one.
        \item We showcase the effectiveness of our approach on a small scale real-world scenario. 
    \end{itemize}
    
    \section{Related Work}
    \label{sec:rel}
    In \cite{jha2002two}, the authors present a formal analysis of attacks on a network with cost-benefit analysis and about potential security measures to defend against the network attacks. In \cite{chowdhary2016sdn}, authors provide a polynomial time method for attack graph construction and network reconfiguration using a parallel computing approach, making it possible to reason about known multi-stage attacks in large-scale systems.
    
    Authors in \cite{jia2013motag} introduced the idea of moving secret proxies to new network locations using a greedy algorithm that can thwart brute force and DDoS attacks. In \cite{zhuang2013investigating}, Zhuang {\em et. al.} show that if intelligent adaptations are used along with these MTD systems, the effectiveness is improved further. In \cite{sengupta2017game} authors show that intelligent strategies based on common intuitions can be detrimental to security and highlight how game theoretic reasoning can alleviate the problem. On those lines, \cite{lye2005game} and \cite{senguptamoving} use a game-theoretic approach to model the attacker-defender interaction as a two-player game where they calculate the optimal response for the players using the Nash and the Stackelberg Equilibrium concepts respectively. Although they propose the use of the Markov Decision Process (MDP) and attack graph-based approaches, they leave it as future work. The flavor of these approaches is similar to those of Stackelberg Security Games (SSGs) that have been used extensively in multiple physical security applications highlighted in \cite{dobss,sinha2015physical}.
    
    In the context of cloud systems, \cite{peng2014moving} discusses a risk-aware MTD strategy where they model the attack surface as a non-decreasing probability density function and then estimate the risk of migrating a VM to a replacement node using probabilistic inference. In \cite{kampanakis2014sdn}, authors highlight obfuscation as a possible MTD strategy in order to deal with attacks like OS fingerprinting and network reconnaissance in the SDN environment. Furthermore, they highlight that the trade-off between such random mutations, which may disrupt any active services, require analysis of cost-benefits.
    
    In this paper, we identify an adaptive MTD strategy against multi-hop monotonic attacks for cloud networks which optimizes for performance while providing gains in security. The ability to decompose a large cloud network into sub-nets provides gains in computing strategies, a fair distribution of IDS resources and prioritizing detection of attacks that may have long-term impacts.
    
    
    Due to lack of solution methods for analyzing the impact of strategically moving the detection surface on multi-stage attacks in cloud systems, we evaluate the effectiveness our defense strategy against two attack strategies-- one static and the other dynamic-- on a simple cloud system. 
    
    
    
    \section{Background}
    \label{sec:background}
    
    In this section, we first introduce our reader to some terminology and the threat model. We then describe a small cloud network scenario that we will use throughout the paper as a representative example to elucidate our ideas.
    \subsubsection{Vulnerability} is a security flaw in a software service hosted over a given port, that when exploited by a malicious attacker, can cause loss of Confidentiality, Availability or Integrity (CIA) of that virtual machine (VM).
    
    \begin{figure}[!t]
        \centering
        \includegraphics[width=0.47\textwidth]{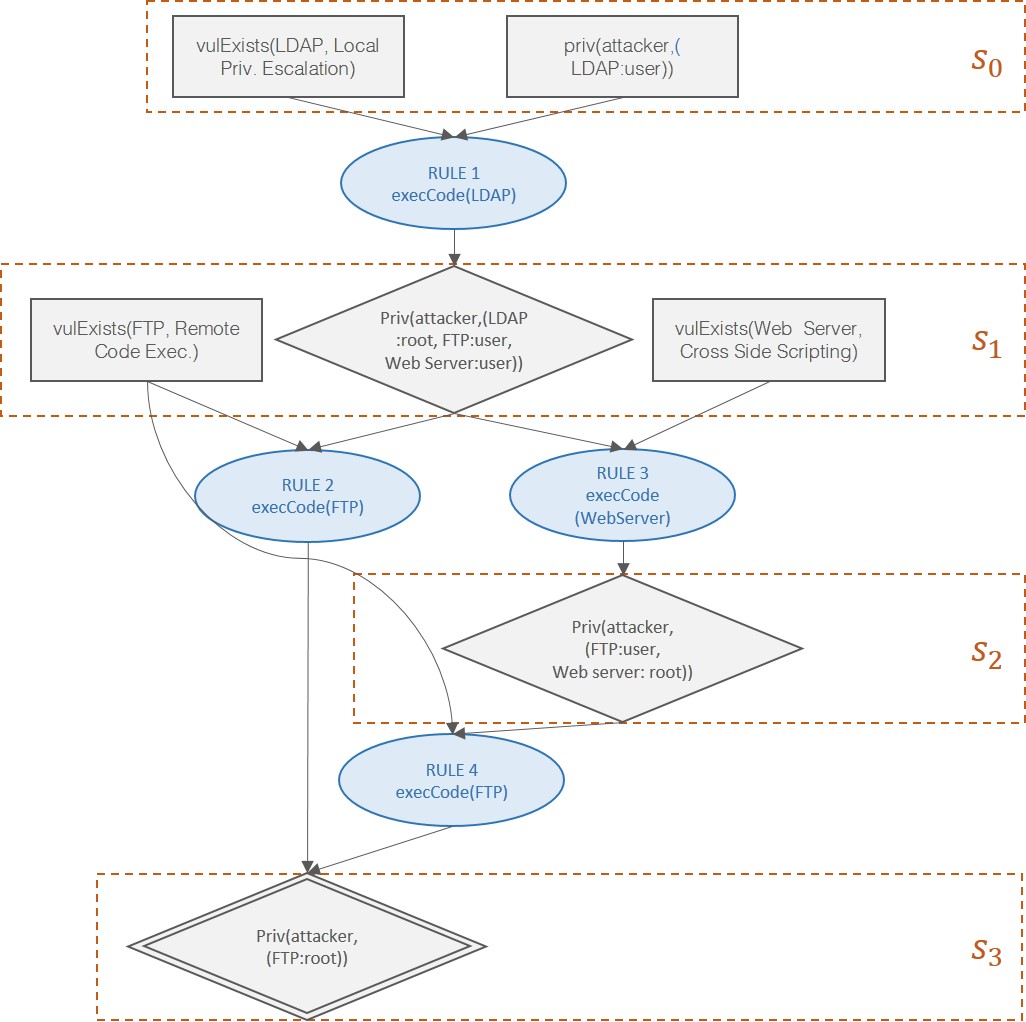}
        \caption{Attack Graph for the example cloud network scenario shown above.}
        \label{fig:scenario1}
        \vspace{-.5em}
    \end{figure}
    
    \subsection{Threat Model}   
    
    \noindent Consider the cloud system in Figure~\ref{fig:scenario_1}, where the attacker has user-level access to the LDAP server, which is the initial state of our game and the goal state is to compromise the FTP server. The attacker can perform actions such as \textit{exploit-LDAP, exploit-Web, exploit-FTP}. In the scenario shown, the attacker has two possible attack paths it can take to reach the goal node \textit{priv(attacker, (FTP: root))}, i.e.
    
    \begin{itemize}
        \item
        Path 1: {\em exploit-LDAP $\rightarrow$ exploit-FTP} 
        \item
        Path 2: {\em exploit-LDAP $\rightarrow$ exploit-Web $\rightarrow$ exploit-FTP}
    \end{itemize}
    
    \noindent The Admin can choose to monitor (1) services running on the host and access to sensitive files in the system, and (2) network traffic along both the paths using the network and host-based monitoring agents, e.g., \textit{monitor-LDAP}, \textit{monitor-FTP}, etc. We assume that the Admin has resource constraints and thus, wants to perform monitoring in an optimized fashion. On the other hand, the attacker should try to perform attacks that have lets it achieve the goal with the highest probability, i.e. avoid being detected by the Admin.
    To model this kind of attack behavior, we utilize the well-established notion of Attack Graphs (AGs).
    \\   
    \subsubsection{Attack Graph} $G=\{N,E\}$ consists of a set of nodes (N) and a set of edges (E) where,
    \begin{itemize}
        \item As shown in the Figure~\ref{fig:scenario1}, the nodes (N) of attack graph can be denoted by $ N = \{ N_f \cup N_c \cup N_d \cup N_r \} $. Here $N_f$ denotes primitive/fact nodes (square boxes in Fig \ref{fig:scenario1}), $N_c$ denotes the exploit (blue circles in Fig \ref{fig:scenario1}), $N_d$ denotes the privilege level (diamond-shaped boxes in Fig. \ref{fig:scenario1}) and $N_r$ represents the root or goal node (diamond-shaped box with double border);
        \item The edges (E) of the attack graph can be denoted by $E =  \{E_{pre} \cup E_{post} \}$. Here $E_{pre} \subseteq (N_f \cup N_c)  \times (N_d \cup N_r)$ ensures that pre-conditions $N_c$ and $N_f$ must be met to achieve $N_d$, i.e., fact (netAccess(VM)) and exploit conditions (vulExists(VM)) should be true for achieving post-condition $N_d$ or $N_r$, e.g., root(VM). $E_{post} \subseteq (N_d \cup N_r) \times (N_f \cup N_c)$ means post-condition $N_d$ or $N_r$ can be achieved on satisfaction of $N_f$ and $N_c$.
    \end{itemize}
    
    \subsection{Two-Player Markov Games}
    
    Having defined the notion of Attack Graphs, we now introduce the concept of Markov Games. Later, we shall see how the information present in attack graphs will be used to define the various aspects of our Markov Game.
    
    \subsubsection{Markov Game} \cite{shapley1953stochastic} for two players $P_1$ and $P_2$ can be defined by the tuple $(S, A_1, A_2, \tau, R, \gamma)$ where,
    \begin{itemize}[nosep]
        \item $S = \{s_1, s_2, s_3, \ldots, s_k\}$ are finite states of the game,
        \item $A_1 = \{a_1^1, a_1^2, \ldots, a_1^m\}$  represents the possible finite action sets for $P_1$,
        \item $A_2 = \{a_2^1, a_2^2, \ldots, a_2^n\}$ are finite action sets for $P_2$,
        \item $\tau(s, a_1, a_2, s')$ is the probability of reaching a state $s' \in S$ for state $s$ if  $P_1$ and $P_2$ take actions $a_1$ and $a_2$ respectively,
        \item  $R^i (s, a_1, a_2)$ is the reward obtained by $P_i$ if in state $s$, $P_i$ and $P_{-i}$ take the actions $a_1$ and $a_2$ respectively, and
        \item $\gamma^i \mapsto [0,1)$ is the discount factor for player $i$. In the rest of the paper, we assume $\forall i ~\gamma^i= \gamma$.
    \end{itemize}
    
    The concept of the optimal policy for a player $P_i$ in this game is defined as the selection of the action that optimizes the value of a being in any state $s$ while reasoning over the expectation of (1) underlying domain stochasticity (defined by $\tau$ and similar to Markov Decision Processes) and (2) reasoning over the other's player $P_{-i}$ action space. This is generally done by finding a min-max policy over the action spaces of both the players in each state, similar to solution strategies in normal (i.e. matrix) or extender form games \cite{littman1994markov}.
    
    Now, notice that in a two-player Markov Game, each state represents a Matrix Game and the policy in each game is not only based on maximizing the reward in this game but also reasoning about the reward to go, which in turn is dependent on the games that you are yet to play. Thus, the max-min strategy seeks to maximize the value for the max player given that the min player selects the pure strategy that gives the minimum pay-off to the max player.  
    To prevent being second-guessed by the min player, the max player should play a mixed strategy, i.e. have a probability distribution over the actions it can play. To formalize this, let us define the Q-values for an action $a_1$ taken by the max player $P_1$ in state $s$, given that $P_2$ selects $a_2$, is defined as,%
    \begin{eqnarray}
        \tiny
        Q(s, a_1, a_2) = R(s, a_1, a_2) + \gamma \sum_{s'} \tau(s, a_1, a_2, s') \cdot V(s')
        \label{eq:Q}
    \end{eqnarray}%
    Let the mixed policy for state $s$ as $\pi(s)$, which is a vector of length $m$ that represents the probability distribution that $P_1$ can has over the possible $m$ actions it can take in state $s$. We can now define the value of state $s$ for $P_1$ using the equation,%
    \begin{eqnarray}
        V(s) = \max_{\pi(s)} \min_{a_2} \sum_{a_1} Q(s, a_1, a_2)\cdot \pi_{a_1}
        \label{eq:V}
    \end{eqnarray}%
    \subsection{Scoring Metrics for Vulnerabilities and Exploits}
    
    \begin{table}[tb]
        \small
        \centering
        \begin{tabular}{ p{0.8cm} | p{2.1cm} | p{1.6cm} | p{0.6cm} | p{1.3cm}}
            \hline
            \textbf{VM} & \textbf{Vulnerability} & \textbf{CVE} & \textbf{CIA} & \textbf{AC}  \\ 
            \hline
            \hline       
            LDAP & Local Privilege Escalation & CVE-2016-5195 & 5.0 & MEDIUM\\ 
            \hline
            Web Server & Cross Site Scripting & CVE-2017-5095 & 7.0 & EASY \\
            \hline
            FTP & Remote Code Execution & CVE-2015-3306 & 10.0 & MEDIUM\\ 
            \hline 
        \end{tabular} 
        \caption{Vulnerability Information for the Cloud Network}
        \label{tab:eg_cve}
        \vspace{-.5em}
    \end{table}
    
    Software security is defined in terms of Confidentiality, Integration, and Availability \cite{mccumber1991information}. In a broad sense, an attack on a web application is defined as a \textit{act} that compromises any of these characteristics. 
    
    In the three VMs shown above-- an LDAP server, an FTP server, and a web server-- the vulnerabilities present in each of them can be mapped to (known) CVE. These vulnerabilities correspond to the attacker's actions, along with a brief description, are shown in Table \ref{tab:eg_cve}.
    
    The use of the Common Vulnerability Scoring System (CVSS) for rating attacks is well studied in security \cite{houmb2010quantifying}. For (most) CVEs listed in the NVD database, we have a six-dimensional CVSS v2 vector,
    which can be decomposed into multiple components that represent Access Complexity (AC), i.e. how difficult it is to exploit a vulnerability, and the impact on Confidentiality, Integrity, and Availability (CIA) gained by exploiting a vulnerability. The values of AC are categorical \texttt{\{EASY, MEDIUM, HIGH\}}, while CIA values are in the range $[0, 10]$ and are shown for each CVE in Table \ref{tab:eg_cve}.
    
    \section{Game Theoretic Modeling}
    
    In this section, we describe, with the help of the example cloud system mentioned above, how the problem of placing detection systems can be formulated as a zero-sum Markov Game.
    The use of a Markov Game model comes with two implicit assumptions-- (1) attacks can be modeled using a Markovian model and (2) both the players have full observability of the state. Beyond these, we assume (1) there is a list of attacks known to both the defender and the attacker (but cannot be fixed either due to lack of manpower or restrictions from third-party who host their code on the cloud network \cite{jajodia2018share,senguptamoving}) and (2) the attacker can be in any node in the system and remain undetected until it attempts to exploit an existing vulnerability, i.e. stealthy attacker \cite{venkatesan2016moving}.
    
    \subsection{States}
    We view the states of our Markov Game as an abstraction over a set of $N_f \cup N_d \cup N_r$ nodes in the Attack Graph (AG). In our example, these correspond to all the grey colored nodes in Figure \ref{fig:scenario1}. A valid abstraction or state-space formulation for our Markov Game must satisfy two properties. First, a grey node in the AG can belong to only one state in the Markov Game. This ensures that the players do not double count an exploit in the system. Second, all the grey nodes in the AG belong to at least one state in the Markov Game. This ensures that no known vulnerability (or attack path to the goal) is missed out. Note that we can choose to consider the entire attack graph as a single state in our Markov Game formulation.
    
    In our example cloud network, we have four states as shown in Figure \ref{fig:scenario1}. The state $s_3$ is a terminal state which has only one action for both the players and gives a high positive reward to the attacker and by the assumption that we have a zero-sum game, a high negative reward to the defender.
    
    
    \subsection{Players and Action Sets}
    The action set for the defender $P_1$ consists of placing an IDS system for detecting specific attacks targeted to exploit the known vulnerabilities in the system. We denote this as, for example, {\tt mon-FTP}, which means that the defender has deployed a {\em snort} like the defense mechanism for monitoring traffic on the FTP port in the particular state. We also use host-based intrusion detection systems like \textit{auditd} for monitoring access to specific files (like {\tt /etc/passwd} which we do not expect to be accessed) on our system. The actions available to the defender in state $s_1$ for our example scenario is shown in Table \ref{r1}.
    
    The action set for the attacker $P_2$ consists of known vulnerabilities present in the cloud system. Although these correspond to exploiting particular CVEs, we use notations, such as {\tt exp-Web} to mean that the attacker uses CVE-2017-5059 to exploit the web server, for simplicity. In the subsection on case-study of a real-world system, we discuss in detail how one can automatically find the known vulnerabilities present in a system. Both the defender and the attacker has to actions in their action set that denote no activity from a particular player (called {\tt no-mon} and {\tt no-op} respectively).
    
    \begin{figure}[!t]
        \centering
        \includegraphics[width=0.48\textwidth]{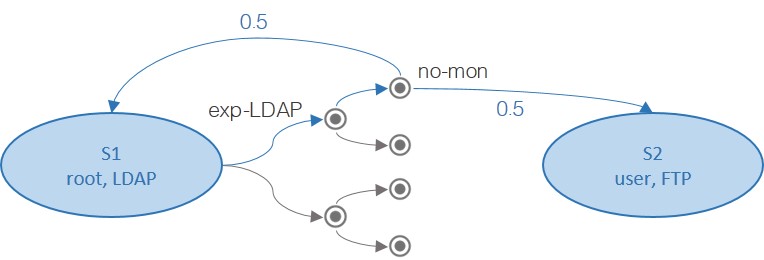}
        \caption{Sample Transition from $s_1$ to $s_2$ when the attacker chooses to exploit an LDAP vulnerability and the defender chooses not to monitor because of resource constraints.}
        \label{fig:TDS}
        \vspace{-.5em}
    \end{figure} 
    
    \subsection{Transitions}
    An example transition in our Markov Game is shown in Figure \ref{fig:TDS}, where players are in state $s_0$. The attacker has root access on an LDAP server (the state) and has two actions available to them-- exploit the vulnerability in the LDAP server or do nothing (for the fear of getting detected). On the other hand, the defender has two actions--either to deploy an IDS that actively monitors an attack on the LDAP server or not monitor at all (due to resource and performance constraints). In the transition shown in Fig \ref{fig:TDS}, when the attacker exploits LDAP and the defender is not monitoring, the attacker has a $50\%$ chance of successfully exploiting it. These probabilities are calculated using the procedure in \cite{chung2013nice} in which they leverage the exploitability scores ($ES$) of each attack which is possible in and from the state from which the transition occurs. In case the attack does not succeed, the players remain in the same state with the remaining probability of $0.5$. Similarly, we have transition values for all the other states and joint actions of the defender and the attacker.

    \subsection{Rewards}
    We consider the rewards for our game to be zero-sum. The reward metrics for each state (except the terminal state) is shown in Table \ref{r1} and \ref{fig:3}. The reward values are obtained using (1) the impact score ($IS$) of a particular attack and (2) the cost of performance degradation based on the placement of a particular IDS at a particular point in the network. Consider Table \ref{r1} and the second row that corresponds to the attacker exploiting the vulnerability {\tt exp-Web}, which in our case maps to the CVE-2017-5059. If the defender does not place IDS to detect attacks on the Web server (first row), it gets a negative reward of $-7$ which is the impact of that vulnerability. If it chooses to deploy the corresponding IDS (second row), the attacker incurs a negative utility of $-5$ and by the virtue of a zero-sum game, the defender gains a reward of $5$ for having stopped the ongoing attack. The reward is short of $7$ because it incurs some performance cost, in this case, worth $2$ utility points. Lastly, if the defender chooses to deploy a monitoring service for detecting exploits on the FTP port, it will not be able to detect an exploit on the web-server and thus incur both the losses for (1) not detecting the attack and (2) having spend resources to deploy an IDS (here worth $3$ utility points).
    
    \begin{table}[t]
        \footnotesize
        \begin{tabular}{cc|c|c|c|}
            & \multicolumn{1}{c}{} & \multicolumn{3}{c}{$P_1$ (Defender)}\\
            & \multicolumn{1}{c}{} & \multicolumn{1}{c}{no-mon}  & \multicolumn{1}{c}{mon-Web} & \multicolumn{1}{c}{mon-FTP} \\\cline{3-5}
            & no-op & $0,0$ & $2,-2$ & $3,-3$ \\\cline{3-5}
            \multirow{1}*{$P_2$ (Atk.)} & exp-Web & $7,-7$ & $-5,5$ & $10,-10$ \\\cline{3-5}
            & exp-FTP & $10,-10$ & $10,-10$ & $-7,7$ \\\cline{3-5}
        \end{tabular}
        \caption{Reward $(R_2, R_1)$ for state $s_1$}
        \label{r1}
        {\tiny
            \begin{tabular}{c|c|c|}
                \multicolumn{1}{c}{} & \multicolumn{1}{c}{no-mon}  & \multicolumn{1}{c}{mon-LDAP} \\\cline{2-3}
                no-op & $0,0$ & $3,-3$ \\\cline{2-3}
                exp-LDAP & $5,-5$ & $-5,5$ \\\cline{2-3}
            \end{tabular}
            \begin{tabular}{c|c|c|}
                \multicolumn{1}{c}{} & \multicolumn{1}{c}{no-mon}  & \multicolumn{1}{c}{mon-FTP} \\\cline{2-3}
                no-op & $0,0$ & $2,-2$ \\\cline{2-3}
                exp-FTP & $10,-10$ & $-8,8$ \\\cline{2-3}
            \end{tabular}
            \vspace{4pt}
            \caption{Reward $(R_2, R_1)$ for states $s_0$ (left) and $s_2$ (right).}
            \label{fig:3}
        }
        \vspace{-.5em}
    \end{table}
    
    \begin{figure*}[t]
    \centering
        \parbox{.45\linewidth}{
            \centering
            \begin{tikzpicture}[]
            \begin{axis}[
            width=\linewidth,
            height=0.75\linewidth,
            xticklabel style={anchor=near xticklabel},
            xlabel={$\gamma \rightarrow$},
            ylabel={Defender's utility},
            grid style={dashed,red},
            font=\scriptsize,
            smooth,
            mark size=1.3pt,
            mark options={solid},
            legend style ={
                at={(0.65,0.05)}, 
                anchor=south east,
                draw=none,
                fill=none,
                font=\scriptsize,
            },
            xtick=data
            ]
            \addplot+[name path=V1_mmp,BrickRed,dotted,mark=o] table[x=gamma,y=V1_mmp] {values.dat};
            \addplot+[name path=V1_ur,NavyBlue,mark=o] table[x=gamma,y=V1_ur] {values.dat};
            \addplot+[name path=V1_om,OliveGreen,dashed,mark=triangle] table[x=gamma,y=V1_om] {values.dat};
            \addlegendentry{Min-Max Pure Strategy}
            \addlegendentry{Uniform Random Strategy}
            \addlegendentry{Optimal Mixed Strategy}
            \end{axis}
            \end{tikzpicture}
        }
        ~~
        \parbox{.45\linewidth}{
            \centering
            \begin{tikzpicture}[]
            \begin{axis}[
            width=\linewidth,
            height=0.75\linewidth,
            xticklabel style={anchor=near xticklabel},
            xlabel={$\gamma \rightarrow$},
            ylabel={Defender's utility},
            grid style={dashed,red},
            font=\scriptsize,
            smooth,
            mark size=1.3pt,
            mark options={solid},
            legend style ={
                at={(0.65,0.05)}, 
                anchor=south east,
                draw=none,
                fill=none,
                font=\scriptsize,
            },
            xtick=data
            ]
            \addplot+[name path=V2_mmp,BrickRed,dotted,mark=o] table[x=gamma,y=V2_mmp] {values.dat};
            \addplot+[name path=V2_ur,NavyBlue,mark=o] table[x=gamma,y=V2_ur] {values.dat};
            \addplot+[name path=V2_om,OliveGreen,dashed,mark=triangle] table[x=gamma,y=V2_om] {values.dat};
            \addlegendentry{Min-Max Pure Strategy}
            \addlegendentry{Uniform Random Strategy}
            \addlegendentry{Optimal Mixed Strategy}
            \end{axis}
            \end{tikzpicture}
        }
        \parbox{.45\linewidth}{
            \centering
            \begin{tikzpicture}[]
            \begin{axis}[
            width=\linewidth,
            height=0.75\linewidth,
            xticklabel style={anchor=near xticklabel},
            xlabel={$\gamma \rightarrow$},
            ylabel={Defender's utility},
            grid style={dashed,red},
            font=\scriptsize,
            smooth,
            mark size=1.3pt,
            mark options={solid},
            legend style ={
                at={(0.65,0.05)}, 
                anchor=south east,
                draw=none,
                fill=none,
                font=\scriptsize,
            },
            xtick=data
            ]
            \addplot+[name path=V3_mmp,BrickRed,dotted,mark=o] table[x=gamma,y=V3_mmp] {values.dat};
            \addplot+[name path=V3_ur,NavyBlue,mark=o] table[x=gamma,y=V3_ur] {values.dat};
            \addplot+[name path=V3_om,OliveGreen,dashed,mark=triangle] table[x=gamma,y=V3_om] {values.dat};
            \addlegendentry{Min-Max Pure Strategy}
            \addlegendentry{Uniform Random Strategy}
            \addlegendentry{Optimal Mixed Strategy}
            \end{axis}
            \end{tikzpicture}
        }
        ~~
        \parbox{.45\linewidth}{
            \centering
            \begin{tikzpicture}[]
            \begin{axis}[
            width=\linewidth,
            height=0.75\linewidth,
            xticklabel style={anchor=near xticklabel},
            xlabel={$\gamma \rightarrow$},
            ylabel={Defender's utility},
            grid style={dashed,red},
            font=\scriptsize,
            smooth,
            mark size=1.3pt,
            mark options={solid},
            legend style ={
                at={(0.65,0.05)}, 
                anchor=south east,
                draw=none,
                fill=none,
                font=\scriptsize,
            },
            xtick=data
            ]
            \addplot+[name path=V0_mmp,BrickRed,dotted,mark=o] table[x=gamma,y=V0_mmp] {values.dat};
            \addplot+[name path=V0_ur,NavyBlue,mark=o] table[x=gamma,y=V0_ur] {values.dat};
            \addplot+[name path=V0_om,OliveGreen,dashed,mark=triangle] table[x=gamma,y=V0_om] {values.dat};
            \addlegendentry{Min-Max Pure Strategy}
            \addlegendentry{Uniform Random Strategy}
            \addlegendentry{Optimal Mixed Strategy}
            \end{axis}
            \end{tikzpicture}
        }
        \caption{Defender's value in each of the four state--$s_0$ (top-left), $s_1$ (top-right), $s_2$ (bottom-left), and $s_3$ (bottom-right).}
        \label{fig:compare}
        \vspace{-.5em}
    \end{figure*}
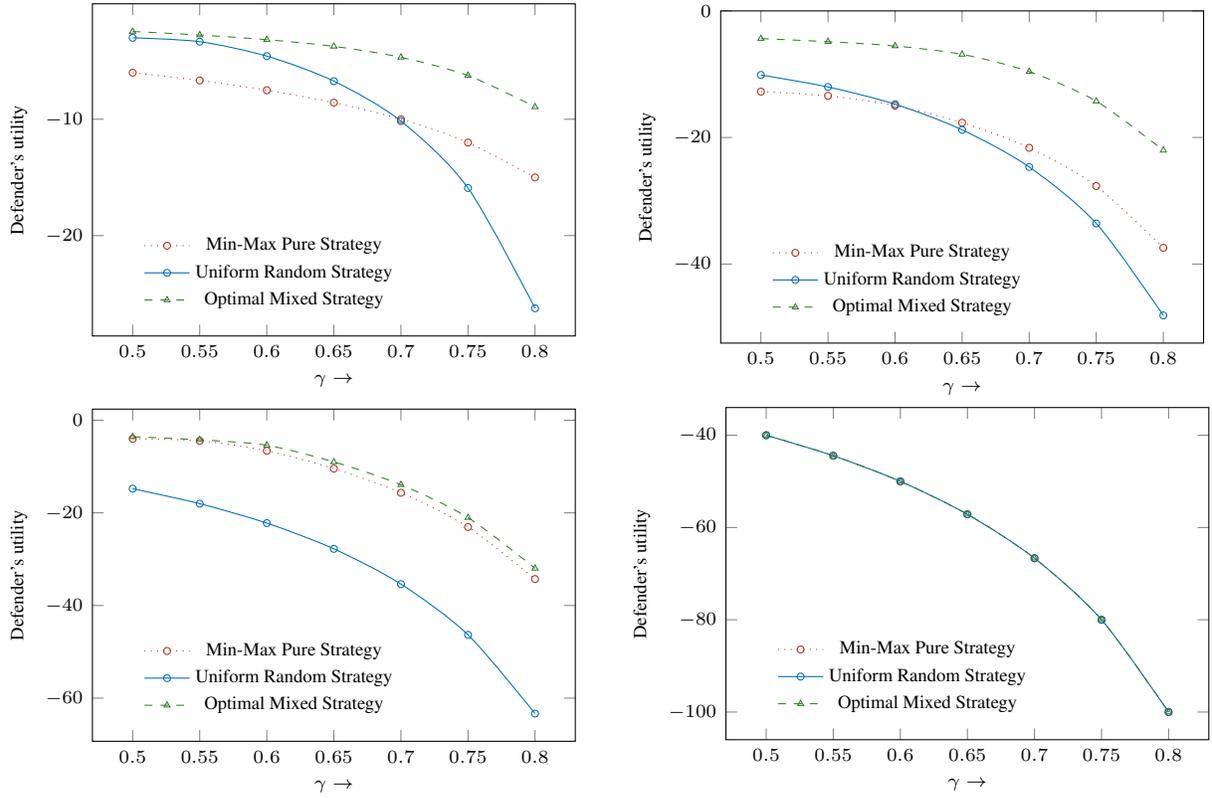
    
    \section{Experimental Results}
    
    In this section, we use the small network shown in Figure \ref{fig:scenario1} to show the effectiveness of the optimal Markov Game strategy against naive baseline methods that are popular in the cybersecurity community.
    
    
    \subsection{Baseline Methods}
    
    \begin{itemize}
        \item \textbf{Min-Max Pure Strategy (MMPS).} The defender selects a pure strategy $a_1$ given that the attacker selects the an action that gives the defender the minimum value. This is similar to the min-max computation we do for fully observable, deterministic games like chess and can be mathematically represented by modifying Equation \ref{eq:V},
        \begin{eqnarray}
            V(s) = \max_{a_1} \min_{a_2} Q(s, a_1, a_2)
            \label{eq:V1}
        \end{eqnarray}%
        where the $\pi(s)$ is replaced with $a_1$. In there exists a pure strategy min-max equilibrium for the Markov Game, i.e. a static placement of IDS that clearly dominates any other placement in regards to security and performance, this would have been the optimal strategy. We do not expect this to happen in real-world scenarios and thus, introduce the notion of Moving Target Defense (MTD) that argues in favor of a mixed strategy that (,as opposed to a pure strategy) makes it harder for the attacker to second guess the defender's move. Having said that, MMPS is the best static placement strategy that a defender can come up under performance constraints and, in most cases, better than what many network administrators use in practice. Thus, MMPS acts as a reasonable baseline.
        
        \item \textbf{Uniform Random Strategy (URS).} In this, the defender uses a uniform probability  distribution over its actions (or pure strategies) in a state. For example, consider state $s_1$ shown in Table \ref{r1}. The defender chooses the mixed strategy of monitoring the FTP server, the web server or none of them, all with the equal probability of $33.33\%$. Thus, in any round, the defender rolls a three-sided fair dice and does whatever comes up. Many researchers had claimed that selecting between what to choose when shifting attack surfaces should be done using a pure (or uniformly) random strategy \cite{zhuang2014towards}. This has been disapproved later by \cite{sengupta2017game}. In this work, we use this as a baseline to reiterate that such strategies based on intuition, as opposed to careful modeling of the problem at hand, can do more harm than good.
    \end{itemize}
    
    \subsection{Preliminary Results with Baselines}
    In Figure \ref{fig:compare}, we plot the utility values in all the four states of our game for both the baseline strategies and the optimal mixed policy for our Markov Game Formulation (obtained using Equation \ref{eq:V}). As the discount factor increases, both the players start valuing future rewards and thereby, select strategies that given them higher value in the long run. At higher values of the discount factor (near $0.85$), the high magnitude of reward in the terminal state $s_3$ affects the values of other states, thereby increasing the magnitude of gain. When the discount factor is small (near $0.5$), the rewards in the future state does not have a substantial impact on the immediate value of a state, thereby reducing the magnitude of gain. In state $s_3$, the URS, the MMPS, and the optimal mixed strategy are all equivalent because it is a terminal state and there is only one action for both the players.
    
    The optimal strategy for the defender for the four different states is as follows for the discount factor $\gamma=0.8$:
    
    \begin{lstlisting}[mathescape=true,
        breaklines=true,
        xleftmargin=\dimexpr-\leftmarginii-\leftmargini
    ]
    $\pi(s_3)$ : {terminate: $1.0$}
    $\pi(s_0)$ : {no-mon: $0.404$, mon-LDAP: $0.596$}
    $\pi(s_1)$ : {no-mon: $0.0$, mon-Web: $0.547$, mon-FTP: $0.453$}
    $\pi(s_2)$ : {no-mon: $0.0$, mon-FTP: $1.0$}
    \end{lstlisting}%
    Note that for states that are closer to the goal ($s_1$ and $s_2$), the defender has $0$ probability for not placing a monitoring system. This means the risk of not monitoring attacks closer to a goal node thereby landing up in the terminal state is much higher than the cost for losing out on performance. On the other hand, for states further away form the goal ($s_0$), the defender places non-zero probability for not monitoring known attacks. In our case, these results indicate that a defender can focus on performance at places near the entry points but should prioritize for security in the states close to the goal.
    
    In our example, the defender can detect only one of the attacks. Although this can be easily addressed by adding more pure strategies to the defender's action set, in the worst case, this may lead to an exponential increase in the size of $|A_1|$. Thus, calculating the min-max strategy becomes computationally expensive. Although we can select states for our Markov Game such that the number of actions in each state is restricted to allow this computation, such abstraction of the Attack Graph may not be practically meaningful. We hope to investigate and address this issue in the future.

\subsubsection{Complexity Analysis} The value function calculated for zero-sum Markov game in Equations (1) and (2) above is guaranteed to converge in polynomial time given that there is a terminal state with high reward for at least one player from which it is impossible to transition into any other state. The value update is more expensive that the value iteration algorithm, in which each iteration takes $O(|S|^2|A|)$ steps because the reasoning happens over the space of joint action space of both the players. Mentioning a tight upper bound is hard because this reasoning involves solving a Linear Program. Given that in the case of cyber security applications, the transition function is sparse (because not all actions are applicable in all state), we can get significant gains in speed. Now, we show a more involved and realistic example on a small-scale cloud network.
    
    \subsection{Case Study: MTD Against Advanced Persistent Threats}
    The attacker performs a multi-stage attack, targeting the services at the gateway of the network first and then trying to penetrate into internal network services. The goal of this attack is to exfiltrate as much information as possible while maintaining persistence over a long period of time. Most attack detection tools just utilize signature-based tools in order to identify the data at the border of the network. Additionally, the tools are configured in an ingress filtering mode, hence the data going out of the network is left unexamined. 
    
    Based on the standards defined by NIST and other organizations~\cite{brewer2014advanced}, the attack analysis from a defender/security administrator’s perspective takes place in five steps, namely: \begin{inparaenum}[\itshape 1)\upshape]
        \item Reconnaissance/ Intelligence Gathering
        \item Threat Modeling
        \item Vulnerability Scanning and Analysis
        \item Exploitation
        \item Post Exploitation
    \end{inparaenum}
    
    \begin{figure}[t]
        \centering
        \includegraphics[width=0.48\textwidth]{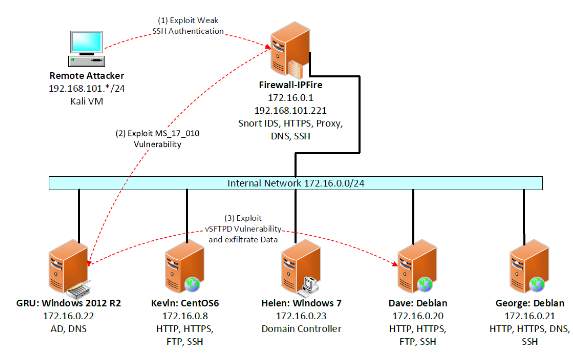}
        \caption{An Advanced Persistent Threat (APT) Scenario}
        \label{fig:apt}
        \vspace{-0.6em}
    \end{figure}
    
    In order to simulate an APT scenario, we created a flat network using the VM images from the Western Region Cybersecurity Defense Competition (WRCCDC)~\cite{wrccdc}. The competition consists of eight Blue Teams from different regions who face a team of experienced hackers (Red Team) from the Industry. The goal of Blue teams is to maintain service availability while ensuring malicious attempts by Red Team members are logged and reported properly. In our experiments, we focus on how effectively we can detect attacks by the Red teams.
    
 
    
    \begin{figure}[t]
        \centering
        \includegraphics[width=0.48\textwidth]{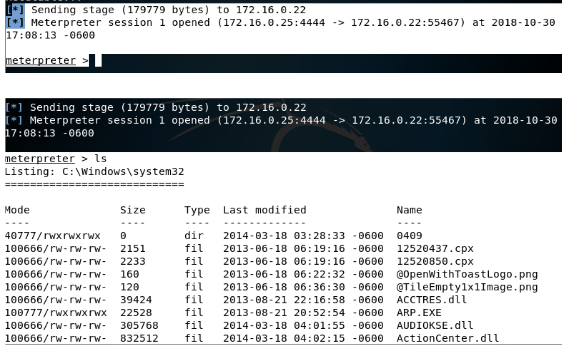}
        \caption{Stage 2 of the APT scenario described above.}
    
        \label{fig:apt2}  
        \vspace{-.5em}
    \end{figure}
    
    We used the VM images from the competition and created a similar environment in ASU's Science DMZ \cite{chowdhary2017science}. We created a flat network with \textit{IPFire (Next-Generation Firewall)} hosted at the gateway of the network (192.168.101.0/24). The VM has the capability to implement traditional Firewall filtering capability. Additionally, the VM has integrated VPN, Snort IDS, Web Proxy for threat detection at different levels of the protocol stack. We now describe the various stages of APT (loosely based on NIST model) arried out by the Red Teams over an extended period of time:
   
\noindent \textbf{Stage1: Slow and Low Weak Authentication Exploit}\quad
    The attacker performs social engineering on website forums frequented by employees of the company. One of the developer's posts a question regarding a key update function for OpenSSH functionality with a specific version (3.3). The attacker identifies this version as being vulnerable to authentication based attack, which can exploit a buffer-overflow vulnerability by sending a well-defined payload to the SSH server hosted at the gateway of the network. In our case, we already knew the vulnerable OpenSSH service.   We consider this as the first step of a multi-stage attack (see Figure \ref{fig:apt}). This represents a scenario how both the players become aware of a known vulnerability present in the system.
  
\begin{table}[!t]
        \small
        \centering
        \begin{tabular}{ p{0.95cm} | p{1.75cm} | p{1.5cm} | p{0.5cm} | p{1.2cm}}
            \hline
            \textbf{VM} & \textbf{Vulnerability} & \textbf{CVE} & \textbf{CIA} & \textbf{AC}  \\ 
            \hline
            \hline
            Firewall & SSH Buffer Overflow & CVE-2017-6542 & 7.5 & MEDIUM\\ 
            \hline
            Win 2012 & Eternal Blue SMB & MS17-010 & 9.3 & HIGH \\
            & Remote Code Execution & MS15-034 & 10.0 & HIGH \\ 
            \hline
            Debian & Anonymous FTP Login & CVE-1999-0497 & 6.4 & MEDIUM\\ 
            \hline 
            Win 7 & MSRPC Service Enumeration & CVE-2008-4250 & 5.0 & MEDIUM\\ 
            & NVT OS End of Life & CVE-2008-4114 & 10.0 & HIGH \\
            
            \hline 
            CentOS 6 & OpenSSL MITM & CVE-2017-3737 & 6.8 & MEDIUM\\ 
            \hline 
        \end{tabular} 
        \caption{Vulnerability Information for the APT scenario.}
        \label{tab:2}
        \vspace{-0.6em}
    \end{table}    
    
\noindent \textbf{Stage 2: Exploiting Windows 7 VM 172.16.0.22} \quad
    The attacker probes the network and identifies the services and OS versions running on the hosts in the network. In our setup, the corporate access control policy allows only Windows systems to interact with resources such as FTP, Web Servers. Thus, the attacker needs to obtain access to a root shell on one of the Windows machines.
    In order to accomplish this, the attacker must target the {\tt MS\_017\_10} vulnerability present on a Windows 2012 R2 server-GRU as shown in Figure~\ref{fig:apt2}, which hosts other services such as Active Directory and Domain Name Server (DNS).
    
\noindent \textbf{Stage 3: Exploiting \textit{vsftpd} vulnerability and exfiltrating data}\quad
    The vsftpd service running on machine Dave has a Debian operating system. The vulnerability on the FTP server can be exploited by the attacker and they can create a backdoor channel to exfiltrate data from FTP server to their command and control center ($C\&C$). Since most organizations have no egress filtering policies for the corporate firewall, so data exfiltration often goes unnoticed. Additionally, the attacker can distribute the data transfer over a period of several weeks even if there is some signature-based rule on IDS to prevent data exfiltration.
    
\noindent \textbf{Stage 4: Post Exploitation}\quad
    The attacker can either use the \textit{meterpreter} (a Kali Linux tool) shell on Windows host to perform privilege escalation and disrupt services if they are a rogue insider or use the windows machine as a jump point for exploiting other machines. The rationale behind exploiting Windows machine first is that Windows acts as a domain controller for many other machines in the network. 
    
    \begin{figure}[t]
        \centering
        \begin{tikzpicture}[]
        \begin{axis}[
        width=\linewidth,
        height=0.7\linewidth,
        xticklabel style={anchor=near xticklabel},
        xlabel={$\gamma \rightarrow$},
        ylabel={Defender's utility},
        grid style={dashed,red},
        font=\scriptsize,
        smooth,
        mark size=1.3pt,
        mark options={solid},
        legend style ={
            at={(0.95,0.05)}, 
            anchor=south east,
            draw=none,
            fill=none,
            font=\scriptsize,
        },
        xtick=data
        ]
        \addplot+[name path=V2_ur,NavyBlue,mark=o] table[x=gamma,y=V2_ur] {real.dat};
        \addplot+[name path=V2_om,OliveGreen,dashed,mark=triangle] table[x=gamma,y=V2_om] {real.dat};
        \addlegendentry{Uniform Random Strategy}
        \addlegendentry{Optimal Mixed Strategy}
        \end{axis}
        \end{tikzpicture}
        \caption{Defender's value in each of the state $s_2$ as discount factor increase. In this state, we consider a sub-system of the entire cloud network in which the defender can place five possible detection systems but chooses to place two out of the five for performance considerations.}
        \label{fig:real}
        \vspace{-1.0em}
    \end{figure}
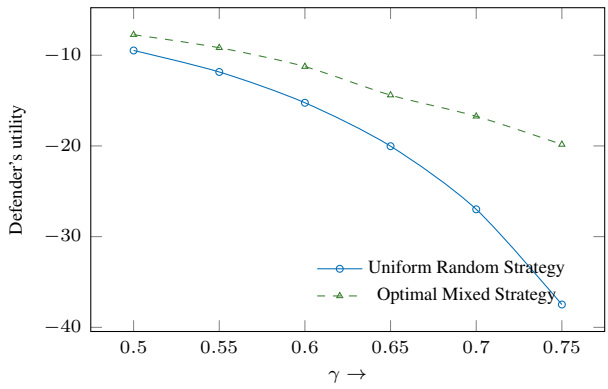
    
    \subsection{Attack Analysis and Results}
    The Blue team identified the following vulnerabilities on the network VMs as shown in the table below. The attacker can have one or more attack goals. One goal of the attack is to ex-filtrate files from the Debian machine (Dave). Another goal can be to target CentOS 6 (Kevin) and disrupt the Domain Name Server (DNS) for the private network. This will in effect lead to service unavailability.    
    
    We show the values a defender, i.e. the Blue Team, obtains if they use an optimal strategy for placement of detection systems for state $s_2$. This state had five possible vulnerabilities and thus, five possible IDS for detecting them (see Fig. \ref{fig:real}). We let the defender provide a limit on the number of IDS systems they can place in this state or sub-net (which was two).
    We saw that, in comparison to the Uniform Random placement strategy in the sub-network represented by state $s_2$, the optimal strategy for the Markov Game yielded better values. Note that our Markov Game formulation treats the number of IDS placed in this subnet independently; regardless of how many IDS systems are placed in other parts of the cloud system. This addresses a major shortcoming of previous research \cite{venkatesan2016moving,senguptamoving} in which some pure strategies can place multiple IDS on the same subnet, thereby affecting its performance, which quality of service in other subnets are not impacted. For APT scenarios, even though the Blue team needs comprehensive logging and monitoring using IDS systems at the granularity of each subnet as well as hosts, monitoring every packet in a cloud network is wasteful in terms of networking and compute resources. We found that the min-max strategy proposed by Markov Game solver in our current work helps in optimizing the number of detection agents while ensuring a high detection rate.    

    \vspace{-0.8em}
    \section{Conclusion and Future Work}
    \label{sec:concl}
    A cloud network is composed of heterogeneous network devices and applications interacting with each other. The interaction of these entities poses both (1) a security risk to overall cloud infrastructure and (2) makes it difficult to secure them. While traditional security solutions provide security mechanisms to detect threats, they fail to reason about multi-stage attacks and at the same time, ignore the performance impact on the cloud system. To address these concerns, we presented a zero-sum Markov Game that provides an intelligent strategy to place detection mechanisms that does maximizes detection of vulnerabilities while considering the performance impact on the cloud network. We show that our methods perform better that static placement mechanisms and Moving Target Defense with Uniform Random Strategy. Lastly, we show how our method can be used for a small scale real-world cloud system.
    
    In the future, we plan to consider the case of general sum games because often, the rewards of the attacker are not exactly opposite to that of the defender \cite{sengupta17aamas}. In such cases, the notion of min-max equilibrium becomes a little more involved. We also plan to relax that assumption that the transition function of the formulated Markov Game is accurately defined by normalizing the Exploitability Scores. Lastly, we also plan to consider the case of zero day attacks, which upon discovery modifies the underlying attack graph and thereby, the formulated Markov Game.
    
\section*{Acknowledgment}
We thank the reviewers for their insightful comments and constructive feedback. This research is supported in part by following research grants: Naval Research Lab N00173-15-G017, AFOSR grant FA9550-18-1-0067, the NASA grant NNX17AD06G, ONR grants N00014-16-1-2892, N00014-18-1-2442, N00014-18-12840, NSF–-US DGE-1723440, OAC-1642031, SaTC-1528099, 1723440 and  NSF–-China 61628201 and 61571375. Sailik Sengupta is supported by the IBM Ph.D. Fellowship.

    
    \balance
    \bibliographystyle{aaai}
    \bibliography{mtd,mtdcns,mtdwa}

\begin{thebibliography}{}

\bibitem[\protect\citeauthoryear{Brewer}{2014}]{brewer2014advanced}
Brewer, R.
\newblock 2014.
\newblock Advanced persistent threats: minimising the damage.
\newblock {\em Network security} 2014(4):5--9.

\bibitem[\protect\citeauthoryear{Chowdhary \bgroup et al\mbox.\egroup
  }{2017}]{chowdhary2017science}
Chowdhary, A.; Dixit, V.~H.; Tiwari, N.; Kyung, S.; Huang, D.; and Ahn, G.-J.
\newblock 2017.
\newblock Science dmz: Sdn based secured cloud testbed.
\newblock In {\em IEEE Conference onNetwork Function Virtualization and
  Software Defined Networks}.

\bibitem[\protect\citeauthoryear{Chowdhary, Pisharody, and
  Huang}{2016}]{chowdhary2016sdn}
Chowdhary, A.; Pisharody, S.; and Huang, D.
\newblock 2016.
\newblock Sdn based scalable mtd solution in cloud network.
\newblock In {\em ACM Workshop on Moving Target Defense}.

\bibitem[\protect\citeauthoryear{Competition}{2018}]{wrccdc}
Competition, W. R. C.~D.
\newblock 2018.
\newblock {WRCCDC}.
\newblock \url{https://archive.wrccdc.org/images/2018/}.

\bibitem[\protect\citeauthoryear{Houmb, Franqueira, and
  Engum}{2010}]{houmb2010quantifying}
Houmb, S.~H.; Franqueira, V.~N.; and Engum, E.~A.
\newblock 2010.
\newblock Quantifying security risk level from cvss estimates of frequency and
  impact.
\newblock {\em JSS} 83(9):1622--1634.

\bibitem[\protect\citeauthoryear{Jajodia \bgroup et al\mbox.\egroup
  }{2018}]{jajodia2018share}
Jajodia, S.; Park, N.; Serra, E.; and Subrahmanian, V.
\newblock 2018.
\newblock Share: A stackelberg honey-based adversarial reasoning engine.
\newblock {\em ACM Transactions on Internet Technology (TOIT)}.

\bibitem[\protect\citeauthoryear{Jha, Sheyner, and Wing}{2002}]{jha2002two}
Jha, S.; Sheyner, O.; and Wing, J.
\newblock 2002.
\newblock Two formal analyses of attack graphs.
\newblock In {\em Computer Security Foundations Workshop, 2002. Proceedings.
  15th IEEE},  49--63.
\newblock IEEE.

\bibitem[\protect\citeauthoryear{Jia, Sun, and Stavrou}{2013}]{jia2013motag}
Jia, Q.; Sun, K.; and Stavrou, A.
\newblock 2013.
\newblock Motag: Moving target defense against internet denial of service
  attacks.
\newblock In {\em 2013 22nd International Conference on Computer Communication
  and Networks},  1--9.
\newblock IEEE.

\bibitem[\protect\citeauthoryear{Kampanakis, Perros, and
  Beyene}{2014}]{kampanakis2014sdn}
Kampanakis, P.; Perros, H.; and Beyene, T.
\newblock 2014.
\newblock Sdn-based solutions for moving target defense network protection.
\newblock In {\em IEEE 15th International Symposium on a World of Wireless,
  Mobile and Multimedia Networks}.
\newblock IEEE.

\bibitem[\protect\citeauthoryear{Littman}{1994}]{littman1994markov}
Littman, M.~L.
\newblock 1994.
\newblock Markov games as a framework for multi-agent reinforcement learning.
\newblock In {\em Eleventh International Conference on Machine Learning}.

\bibitem[\protect\citeauthoryear{Lye and Wing}{2005}]{lye2005game}
Lye, K.-W., and Wing, J.~M.
\newblock 2005.
\newblock Game strategies in network security.
\newblock {\em International Journal of Information Security}.

\bibitem[\protect\citeauthoryear{McCumber}{1991}]{mccumber1991information}
McCumber, J.
\newblock 1991.
\newblock Information systems security: A comprehensive model.
\newblock In {\em Proceedings of the 14th National Computer Security
  Conference}.

\bibitem[\protect\citeauthoryear{Paruchuri \bgroup et al\mbox.\egroup
  }{2008}]{dobss}
Paruchuri, P.; Pearce, J.~P.; Marecki, J.; Tambe, M.; Ordonez, F.; and Kraus,
  S.
\newblock 2008.
\newblock Playing games for security: An efficient exact algorithm for solving
  bayesian stackelberg games.
\newblock In {\em AAMAS, 2008},  895--902.

\bibitem[\protect\citeauthoryear{Peng \bgroup et al\mbox.\egroup
  }{2014}]{peng2014moving}
Peng, W.; Li, F.; Huang, C.-T.; and Zou, X.
\newblock 2014.
\newblock A moving-target defense strategy for cloud-based services with
  heterogeneous and dynamic attack surfaces.
\newblock In {\em IEEE International Conference on Communications (ICC)}.

\bibitem[\protect\citeauthoryear{Sengupta \bgroup et al\mbox.\egroup
  }{2017a}]{sengupta2017game}
Sengupta, S.; Vadlamudi, S.~G.; Kambhampati, S.; Doup{\'e}, A.; Zhao, Z.;
  Taguinod, M.; and Ahn, G.-J.
\newblock 2017a.
\newblock A game theoretic approach to strategy generation for moving target
  defense in web applications.
\newblock AAMAS.

\bibitem[\protect\citeauthoryear{Sengupta \bgroup et al\mbox.\egroup
  }{2017b}]{sengupta17aamas}
Sengupta, S.; Vadlamudi, S.~G.; Kambhampati, S.; Zhao, Z.; Doup{\'e}, A.;
  Taguinod, M.; and Ahn, G.-J.
\newblock 2017b.
\newblock A game theoretic approach to strategy generation for moving target
  defense in web applications.
\newblock In {\em AAMAS}.

\bibitem[\protect\citeauthoryear{Sengupta \bgroup et al\mbox.\egroup
  }{2018}]{senguptamoving}
Sengupta, S.; Chowdhary, A.; Huang, D.; and Kambhampati, S.
\newblock 2018.
\newblock Moving target defense for the placement of intrusion detection
  systems in the cloud.
\newblock {\em Conference on Decision and Game Theory for Security}.

\bibitem[\protect\citeauthoryear{Shapley}{1953}]{shapley1953stochastic}
Shapley, L.~S.
\newblock 1953.
\newblock Stochastic games.
\newblock {\em Proceedings of the national academy of sciences}
  39(10):1095--1100.

\bibitem[\protect\citeauthoryear{Sinha \bgroup et al\mbox.\egroup
  }{2015}]{sinha2015physical}
Sinha, A.; Nguyen, T.~H.; Kar, D.; Brown, M.; Tambe, M.; and Jiang, A.~X.
\newblock 2015.
\newblock From physical security to cybersecurity.
\newblock {\em Journal of Cybersecurity} 1(1):19--35.

\bibitem[\protect\citeauthoryear{Venkatesan \bgroup et al\mbox.\egroup
  }{2016}]{venkatesan2016moving}
Venkatesan, S.; Albanese, M.; Cybenko, G.; and Jajodia, S.
\newblock 2016.
\newblock A moving target defense approach to disrupting stealthy botnets.
\newblock In {\em Proceedings of the 2016 ACM Workshop on Moving Target
  Defense},  37--46.
\newblock ACM.

\bibitem[\protect\citeauthoryear{Zhuang \bgroup et al\mbox.\egroup
  }{2013}]{zhuang2013investigating}
Zhuang, R.; Zhang, S.; Bardas, A.; DeLoach, S.~A.; Ou, X.; and Singhal, A.
\newblock 2013.
\newblock Investigating the application of moving target defenses to network
  security.
\newblock In {\em 6th International Symposium on Resilient Control Systems
  (ISRCS)}.
\newblock IEEE.

\bibitem[\protect\citeauthoryear{Zhuang, DeLoach, and
  Ou}{2014}]{zhuang2014towards}
Zhuang, R.; DeLoach, S.~A.; and Ou, X.
\newblock 2014.
\newblock Towards a theory of moving target defense.
\newblock In {\em Proceedings of the First ACM Workshop on Moving Target
  Defense},  31--40.
\newblock ACM.

\end{thebibliography}

\end{document}